\documentclass[twocolumn]{article}
\usepackage[utf8]{inputenc}
\usepackage{graphicx}
\usepackage[backend=bibtex]{biblatex}
\usepackage{geometry}
\usepackage{subcaption}
\usepackage{colortbl}
\usepackage[dvipsnames]{xcolor}
\usepackage{amsmath}
\usepackage{algorithmic}
\usepackage[backend=bibtex]{biblatex}

\definecolor{lightgray}{rgb}{0.9,0.9,0.9}

\title{Bidirectional American Sign Language to English Translator}
\author{Hardie Cate\\ \texttt{ccate@stanford.edu} \and Zeshan Hussain\\ \texttt{zeshanmh@stanford.edu}}
\date{December 11, 2015}

\begin{document}

\maketitle

\section{Introduction}
In the US alone, there are approximately 900,000 hearing-impaired people whose primary mode of conversation is sign language. 
For these people, communication with non-signers is a daily struggle, and they are often disadvantaged when it comes to finding a job, accessing health care, etc. There are a few emerging technologies designed to translate sign language to English in real time, but most of the current research attempts to convert raw signs into English words. This aspect of the translation is certainly necessary, but it does not take into account the grammatical differences between signed languages and spoken languages. In this paper, we outline our bidirectional translation system that converts sentences from American Sign Language (ASL) to English, and vice versa. 

To perform machine translation between ASL and English, we utilize a generative approach. Specifically, we employ an adjustment to the IBM word-alignment model 1 (IBM WAM1)\cite{ibm} where we define language models for English and ASL, as well as a translation model, and attempt to generate a translation that maximizes the posterior distribution defined by these models. Using these models, we are able to quantify the concepts of fluency and faithfulness of a translation between languages. 

\section{Related Work}

In examining papers and projects related to this topic, we have discovered that translation between signed and spoken languages is still largely an open problem. The few emerging technologies that attempt to tackle this issue are only capable of translating single words or short phrases of ASL into English. In fact, most approaches focus on analyzing videos of signs and converting these into words, which a number of recent CS229 projects have done. 
However, several teams at the University of Pennsylvania have taken a more grammatical approach \cite{zhao2000machine} to the problem of English-to-ASL translation but not in the other direction. There has also been an Italian team that has designed a tentative translation system for Italian sign language,\cite{mazzei2013deep} but there are very few parallels with respect to grammatical structures that we can draw for this project. In short, there has been relatively little emphasis in the literature on the conversion between the grammars of the languages, which is what we examine in this project.

\section{Task Definition}
Let us formalize the notation for this task. For the sake of simplicity, we will use ASL as our “source language” and English as our “target language." In general, $S$ will be used to refer to the sentence in ASL, where $S$ is a sequence of signs $S_1,\ldots,S_m$ and $m$ is the length of the sentence. Each sign $S_i$ is represented by a word with uppercase letters. We also include special "gesture tokens," which represent gestures or hand movements (e.g., [point], [head\_shake]) that are not signs in and of themselves but rather modify other signs in the sentence. As a special case, which we address later, commas are also treated as individual sign tokens. Similarly, we will use $E$ to refer to the English sentence, where $E$ is a sequence of words $E_1,\ldots,E_l$ and $l$ is the length of the sentence.
\begin{align*}
S_{input} &: \mathrm{WRISTWATCH\ [point],\ WHO\ GIVE-YOU?}\\
E_{output} &: \mathrm{Who\ gave\ you\ that\ wristwatch?}
\end{align*}
In the backward direction, the input is a grammatically correct English sentence and the output is a sequence of signs represented by words. An example for this input is 
\begin{align*}
E_{input} &: \mathrm{He\ fell\ down.}\\
S_{output} &: \mathrm{FELL.}
\end{align*}
To evaluate translations generated by our system as well as the baselines in both directions, we use the BLEU-2 score,\cite{papineni2002bleu} which takes in a predicted sentence and a reference sentence, both in the target language.   
Let $p_n$ be the the ratio of the number of shared n-grams to the total number of n-grams in the predicted sentence.
Then for a predicted sentence $p$ and reference sentence $s$, the BLEU-N score is given by
\begin{align*}
BLEU2(p,r)=e^{(1-\frac{|r|}{|p|})}\prod_{i=1}^Np_N.
\end{align*}
Thus BLEU-2 is essentially the ratio of shared unigrams and bigrams between the predicted translation and the gold-standard translation to the number that appear in the predicted translation.
The exponential factor at the beginning of the above expression is called the brevity penalty.
Because the rest of the BLEU expression measures the precision of the prediction, it does not penalize the prediction for leaving out words.
The role of the brevity penalty is to penalize predicted sentences that are shorter than their reference counterparts. As a concrete example, consider translating $\mathrm{"YOU\ LIKE\ COLOR\ BROWN?"}$ to the English $\mathrm{"You\ like\ brown\ color?"}$ with the correct reference $\mathrm{"Do\ you\ like\ the\ color\ brown?"}$.
The BLEU-2 score of this prediction is $e^{1-\frac{6}{4}}(\frac{4}{4})(\frac{1}{3})=0.202$

The BLEU-2 metric is used primarily to rate translations at a corpus level, so it does not perform as well when used to evaluate translations of single sentences. Despite that, the BLEU-2 score should be a sufficient evaluation metric for the purposes of this project, since it is still the most common machine translation evaluator and is relatively simple to interpret.

\section{Data}

Our data consists of 579 ASL/English translation pairs scraped from a repository from lifeprint.com,\cite{aslcorpus} a website dedicated to ASL education. Each pair is a single accurate translation of an ASL sentence or question into English. Many of the ASL sentences contain gesture tokens. These pairs are not specifically designed to be translations from English to ASL and in general, translation between languages is not symmetric. However, we make the simplifying assumption that this translation is symmetric primarily because we do not have a thorough enough understanding of ASL to translate from English to ASL.

To test our algorithms, we split the data into two main sets--a training set (80\% of original data) and a testing set (remaining 20\%). We also have a third set, called the development set, that we use to tune our hyperparameters. This set is constructed by pulling out 10 examples from the training set and treating them as our development set. A more standard convention is to use a 70\%-30\% split for training and testing data, but we decided to use a larger training set size because we use these signs as our entire corpus for our ASL language model. We also use small subsets of the data to run tests to assess the performance of specific aspects of our system. For example, we construct a test set from the original test set to judge the performance of "comma-trigram" feature of our ASL language model. We construct another test set to handle gesture tokens. To assess these tests, we use the BLEU-2 metric described in the task definition.

\section{Technical Approach}
Below, we detail the baselines, oracles, and the IBM word alignment method for this problem. 

\subsection{Baseline and Oracle} 
Because of the bidirectional nature of the problem, we have two baselines. For a baseline in English-to-ASL direction, we use a unigram cost function over English to detect the “most important” words in the sentence and directly translate each word into ASL, preserving the order of these words. To pick the most important words, our baseline selects the highest cost words above some threshold based on our unigram cost function. In this case, a higher cost implies a lower frequency in the English language, suggesting higher importance. For our baseline in the other direction, we directly translate each sign into English and insert small helper words (such as “a,” “the,” “and,” etc.), treating this as a search problem that tries to minimize the cost of the English sentence using a bigram cost function. To test these baselines, we utilized the same BLEU score as we did for the results generated by our IBM word alignment system. This approach allowed us to easily compare the baseline results and IBM WAM results, which are reported in the 'Results' section. 

For our oracles in both directions, we have someone familiar with both languages translate each sentence into the other language. These translations are adequate oracles because they make the most sense from a human standpoint. Furthermore, the oracles serve as appropriate gold standard translations that can be used in the BLEU score calculation.

\subsection{Custom WAM}
The IBM WAM1 is a common machine translation model that we customize to take advantage of common constructions in ASL and allow us to modify how the translation model and language models are constructed. The custom model is meant to give us flexibility in our translations.

We now describe the general workflow of our model in the ASL to English direction. This workflow applies to the other direction as well, except with $E$ switched with $S$: 1). For each potential translation, we calculate $p(E)$ 2). Then, given this channel input, we calculate $p(S|E)$ in the “noisy channel” 3). Finally, we choose the $E$ that gives the highest posterior probability given these two probabilities (likelihood and prior). The $p(E)$ is represented by the language model while the $p(S|E)$ is represented by the translation model. We will derive this formulation below. 

\subsubsection{Modeling}
For concreteness, we will use a running translation example in the descriptions of our models and algorithms. Consider the following gold-standard translation from ASL to English:
\begin{align*}
S_{input} &: \mathrm{YOU\ LIKE\ LEARN\ SIGN?}\\
E_{output} &: \mathrm{Do\ you\ like\ learning\ sign\ language?}
\end{align*}
To derive the models that we will use in the translation, we consider the following optimization problem (note that once again we are writing this derivation in the ASL to English direction but the same derivation applies to the other direction):

Given a sentence S in ASL, we want to find an English sentence such that:
\begin{align*}
E &= \arg\max_{E \in English} P(E|S)\\
E &= \arg\max_{E \in English} P(S|E)P(E)
\end{align*}
The probability $P(E)$ represents the fluency of $E$ in English, i.e. how much the translation makes sense to a native speaker, while $P(S|E)$ represents the faithfulness of the translation from $S$ to $E$, i.e. if the translation reflects the actual meaning of what is being said. We determine $P(E)$ using a language model for English and $P(S|E)$ using a translation model from English to ASL. (Note that in our overall translation from ASL to English, we model the probability of a given translation from English to ASL.) 

\subsubsection{Language Model}
To compute $P(E)$ for a given English sentence $E$, we use an $n$-gram cost function. For instance, in our example sentence $E$ above in the case where $n=3$,
\begin{align*}
p(E) &= p_3(\mathrm{null,null,"do"}) \cdot p_3(\mathrm{null,"do","you"})\\
&\cdot p_3(\mathrm{"learning","sign","language?"})
\end{align*}
where $p_3$ is a 3-gram cost function. We have imported $n$-gram cost functions from http://www.ngrams.info/ \cite{ngrams} for $n=2,3,4,5$.

There are a couple of caveats with the above approach when trying to create a language model for ASL, which is necessary for translation in the other direction (i.e. to calculate $P(S)$). The first is that our ASL language model is basically a unigram model that has additional components. In addition to each sign having an associated probability, which is a simple unigram model, we add a comma trigram construction. This construction incorporates into the language model the following grammatical structure that is commonly found in ASL, 
\begin{align*}
\mathrm{([NOUN]|[NOUN_{PHRASE}])[,][PHRASE]}
\end{align*}
In other words, an infrequent (uncommon) noun will be followed by a comma, which is followed by a phrase that generally describes or refers to the noun. The phrase after the comma usually starts with a more frequent (common) word. The way the comma trigram encapsulates this is by increasing the unigram probability of the second word and decreasing the unigram probability of the first word. This effectively favors a big difference between the word before the comma and the word after the comma, which in turn favors an uncommon word being translated before the comma and a common word after the comma. 
The second caveat is that we incorporate gesture tokens, which are contained in brackets, into our corpus. Our language model does not, however, treat these gesture tokens any differently than regular signs.  

\subsubsection{Translation Model}
Now, we wish to estimate the probability $P(S|E)$ given a trained translation model. We represent this model as a mapping from "sign-English" pairs to probabilities (of those pairs) using a dictionary in python. Training this model requires us to introduce a set $A=a_1,\ldots,a_J$ of alignment variables where $a_j$ represents the index of the word in the English sentence that translates to the $j$th word $s_j$ in $S$. We also allow these variables to take on a value of 0, which represents a “null” word in the English sentence. This null word allows for the possibility that there is no word in the English sentence that directly translates to some sign $s_j$ in $S$. In the case of our example above, if $A$ specifies that “learning” in $E$ translates to “LEARN” in $S$, then $a_3=4$, since “LEARN” is the third word in $S$ and “learning” is in the fourth position of $E$. 
Thus, $A$ can be thought of as a many-to-one mapping from words in $E$ to signs in $S$. (In general, this mapping would need to be many-to-many since we sometimes require phrase-to-phrase translation, but for our purposes the many-to-one assumption is reasonable because generally speaking, multiple English words map to only one sign.) Then, to compute $P(S|E)$, we marginalize out the alignment variables
$$\sum_AP(S,A|E) = \sum_AP(S|A,E)P(A|E).$$ 
Now, let $I$ be the number of words in $E$. For fixed $I$, we assume that each possible alignment for each length is equally likely, which gives us that $P(A|E)=\frac{\epsilon}{(1+I)^J}$. This probability can be thought of as a normalization factor for the probability, $P(S|E)$. Indeed, $P(S|A,E)$ is given by, 
\begin{align*}
p(S|A,E) &= \prod_{j=1}^J t(s_j,e_{a_j}),
\end{align*}
where $t(s,e)$ is the probability of translating $e$ as $s$. Therefore, this gives us the following for $P(S|E)$, 
\begin{align*}
p(S|E) &= \sum_A\frac{\epsilon}{(1+I)^J}\prod_{j=1}^J t(s_j,e_{a_j}),
\end{align*}
After training on our dataset, constructing a language model for each language using the corpa we have processed, and generating a parameterized translation model using the EM algorithm, we can use our decoder to find the optimal translation. Both the EM and decoding algorithms are described in detail in the next section. 

\subsubsection{Algorithms}

The two primary algorithms used in our implementation are an EM algorithm to train our translation model which defines $P(S|E)$ and our decoding algorithm. The decoding algorithm is a modified version of the decoding in IBM WAM1, which uses a variant of beam search to find
\begin{align*}
\arg\max_E P(S|E)P(E).
\end{align*}
First we describe the EM algorithm. Our language model maintains a mapping $t$ from pairs $(s, e)$, where $s$ is a sign and $e$ is an English word, to probabilities $t(s,e)=p(s|e)$. Given our training data containing a list of $m$ translation pairs $(S, E)$, we can rewrite our estimate of a single transition probability as
\begin{align*}
p(s|e)&=\sum_{i=1}^m\sum_{A_i}p(S_i,A_i|E_i)\\
&=\sum_{i=1}^m\sum_{A_i}p(A_i|E_i)\prod_{j=1}^J t(s_j,e_{a_j})
\end{align*}
where $A_i$ is the alignment vector for training example $i$. To estimate $t(s,e)$ given the data, we need to find the maximum likelihood of this probability. However, with the introduction of these latent variables,  it is not possible to find this likelihood in a closed form. Therefore, we use the EM algorithm to estimate this likelihood by computing the probability of each possible alignment given a distribution of $t(s,e)$ for each $s$ and $e$ in the E-step and then adjusting this distribution based on these alignment probability estimates in the M-step. The algorithm runs as follows:
\begin{algorithmic}
\STATE{\textbf{EM for ASL-to-English translation model}}
\WHILE{Not converged}
    \STATE{Set $t(s,e)$ uniform, including $t(s,\textnormal{NULL})$.}
    \STATE{Set counts of $e$ translating to $s$ as $C(s,e)=0 \forall s,e$.}
    \FOR{$(S,E)$ in training set}
        \FOR{sign position $i=1,\dots, l$}
            \STATE{$P(a_i=j|S,E)=\frac{t(s_i,e_j)}{\sum_{j'}t(s_i,e_{j'})}$}
            \STATE{$C(s_i|e_{a_i}) += P(a_i=j|S,E)$}
        \ENDFOR
    \ENDFOR
    
    \FOR{all $s,e$}
        \STATE{$t(s,e)=\frac{C(s,e)}{\sum_{s'}C(s',e)}$}
    \ENDFOR
\ENDWHILE
\end{algorithmic}
Our decoding algorithm is a modification of the conventional decoding algorithm used in IBM WAM1. We frame the problem of finding the best English translation as a type of search problem. Let $l$ be the length of $S$. The algorithm is a variant on BEAM search that maintains a list of priority queues $q_0,q_1,\dots,q_l$ of hypotheses. Each hypothesis consists of a list of English words $e$ each with a corresponding sign $s_e$ from $S$. Initially, all priority queues are empty, with the exception of $q_0$, which contains the empty hypothesis. Then we iterate through the priority queues, and for the top hypothesis $h$ in the current queue, we generate some number of new hypotheses $h_1,\dots,h_k$ by adding a single English word as the next word in $h$. We place each new hypothesis $h'$ in $q_i$, where $i$ is the number of signs in $S$ that have been translated to English words in $h'$. Each hypotheses $h$ is prioritized according to
\begin{align*}
\log\sum_{j=1}^{|E|}p(s_{e_j}|e_j)+W\log p(E)
\end{align*}
where $E=[e_1\:e_2\:\dots e_{|E|}]$ is the English sentence contained in $h$ and W is the language model weight whose purpose is explained below. The aspect related to BEAM search is that we only consider the first $k$ hypotheses in each $q_i$ before moving on to $q_{i+1}$. Thus, we effectively prune hypotheses that do not appear among the first $k$ for a given number of translated signs.

\begin{algorithmic}
\STATE{\textbf{Modified decoding algorithm}}
\STATE{Set $q_0$ to PQ with empty hyp.} 
\STATE{Set each $q_1,\dots,q_l$ to empty PQ}
\FOR{$i=0$ to $l-1$}
    \FOR{$j=0$ to MAX\_QUEUE\_SIZE}
        \STATE{hyp = $q_i$.pop()}
        \FOR{$k=1$ to $l$}
            \STATE{newhyp = max over $s,e$ of newhyp($s,e$)}
            \STATE{insert new\_hyp into appropriate queue} 
        \ENDFOR
    \ENDFOR
\ENDFOR
\STATE{Return $q_l$.pop()}
\end{algorithmic}
where newhyp($s,e$) is the new hypothesis formed by translating $s\in S$ to $e$ and newhyp is the maximum according to the priority measure. Once the algorithm reaches the final priority queue, all $l$ signs have been translated. This implies that the top hypothesis in the priority queue is the hypothesis that translates all signs in $S$ to English (possibly to the NULL word) and maximizes $p(S|E)p(E)$, i.e., the hypothesis with English sentence
\begin{align*}
E=\arg\max_{e\in q_l}p(e|S).
\end{align*}

\subsubsection{Algorithms Commentary}
One of the most important aspects of the algorithm is that a hypothesis $h$ can generate a new hypothesis $h'$ that is placed in the same queue as $h$ if it is generated by translating a sign already translated in $h$.
This is important because it allows us to translate a single sign to multiple English words. 
In particular, this allows the translated English sentence to be longer than the input sign sentence, which is the case with most English translations.

Another important note concerns the language model weight. 
If this weight is set to 1, the priority is simply $p(E|S)$, i.e., the value we are trying maximize.
However, we find that if we try to maximize this value, the language model probabilities "outweigh" the probabilities of the translation model.
In effect, the sign sentence is likely to be translated as a set of common English n-grams unrelated to the original sentence.
For instance, when we attempt to translate $\mathrm{YOUR\ SISTER\ SINGLE?}$ as $\mathrm{"Is\ your\ sister\ single?"}$, our top result is $\mathrm{"in\ the\ first"}$.
However, when we introduce this language model weight with $W=0.3$, we produce the translation $\mathrm{"Your\ sister\ single?"}$.
Although this translation might not be as common in English as $\mathrm{"in\ the\ first"}$, it certainly preserves the meaning of the original sentence much better.

Just as with BEAM search, our decoding algorithm is not guaranteed to converge to the optimal result (in fact, this problem in NP-complete), but in many cases it provides good results.

\section{Results}
\begin{figure}[t]
\centering
    \includegraphics[width=1.0\linewidth]{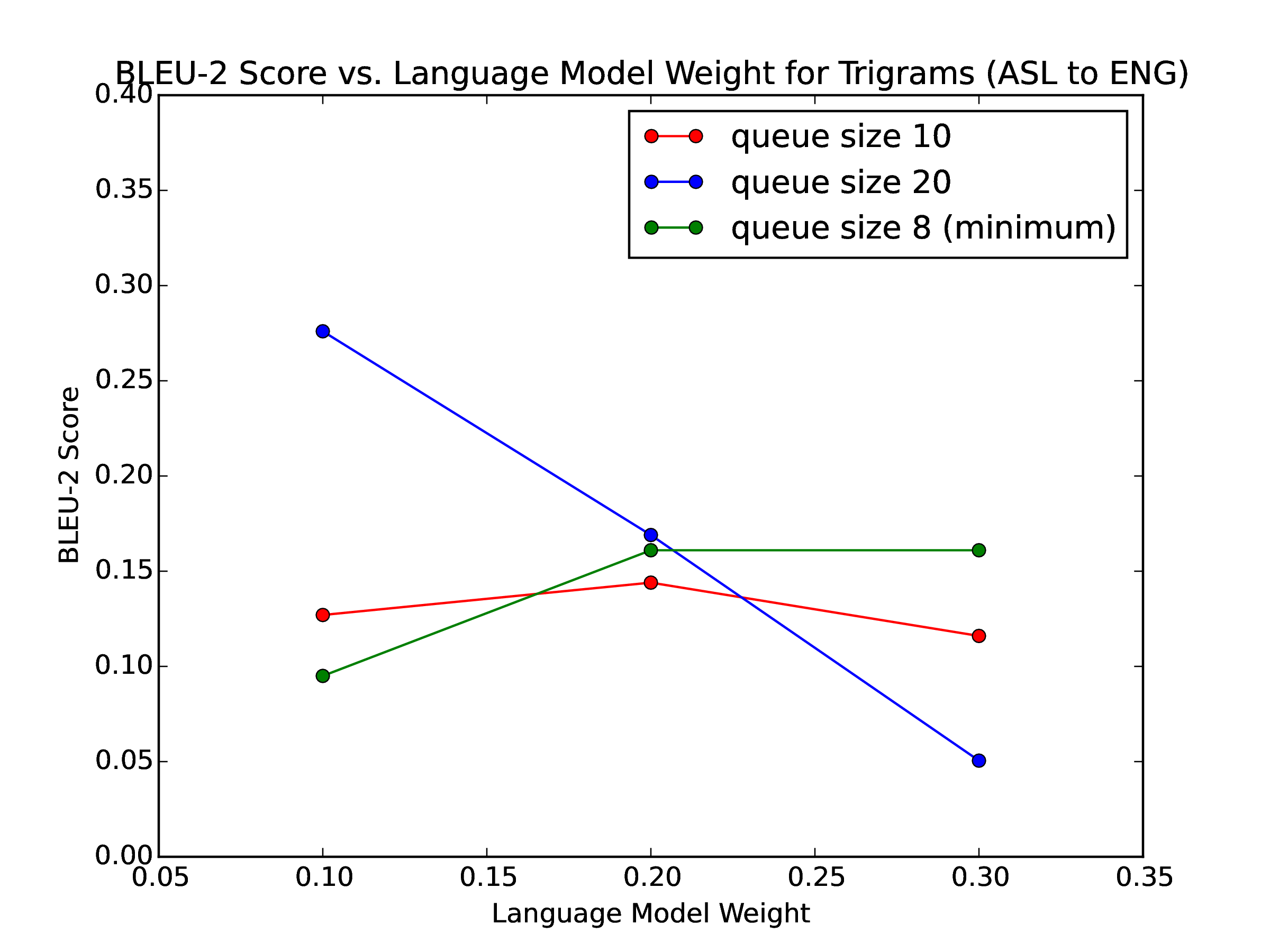}
    \caption{ASL-to-ENG: Varying Language Model Weight}
\label{}
\end{figure}
\begin{figure}[t]
\centering
    \includegraphics[width=1.0\linewidth]{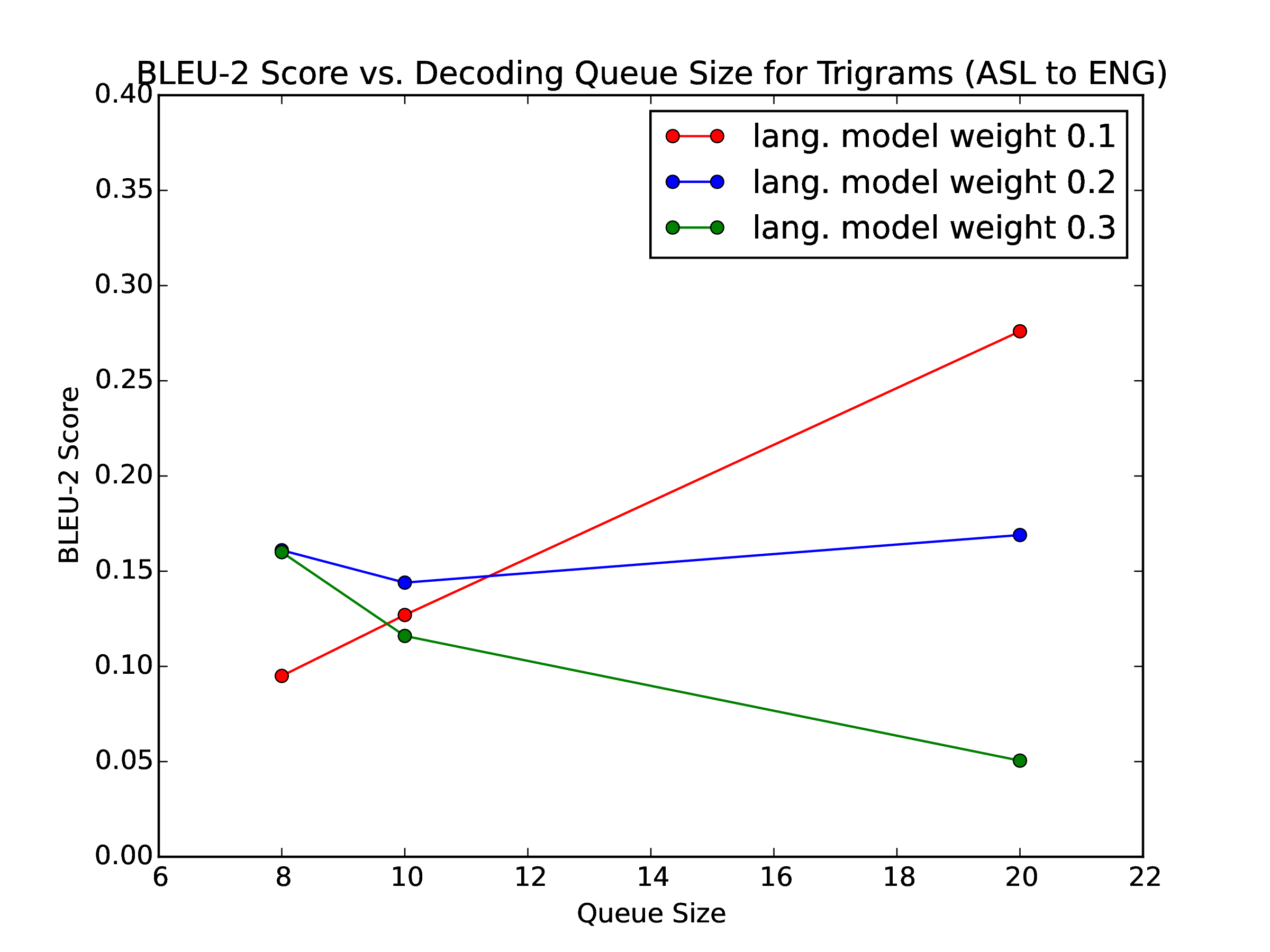}
    \caption{ASL-to-ENG: Varying Queue Size}
\label{fig:my_label}
\end{figure}
\begin{table}[b]
    \centering
    \caption{Overall BLEU-2 Scores on Test Sets}
        \begin{tabular}{|c|c|}
        \hline
        BLEU-2 Score & \\\hline
        ASL-to-English & $0.1202$  \\
        English-to-ASL & $0.1802$ \\
        Comma Trigram Test & $0.1201$ \\ 
        Gestures Test & $0.1232$  \\
        Baseline ASL-to-English & $0.1012$ \\ 
        Baseline English-to-ASL & $0.1139$ \\\hline
        \end{tabular}
    \label{table:comparison}
\end{table}

\begin{table}[b]
    \centering
    \caption{ASL-to-ENG: Bigrams}
    \resizebox{0.95\columnwidth}{!} {
        \begin{tabular}{|c|c|c|}
        \hline
        Queue Size & Language Model Weight & BLEU-2 Score\\\hline
            8 (bigram) & 0.1 & 0.2360 \\\
             & 0.2 & 0.1690 \\
             & 0.3 & 0.1600 \\\hline
            10 (bigram) & 0.1 & 0.1990 \\
             & 0.2 & 0.1690 \\
             & 0.3 & 0.0700 \\\hline
            20 (bigram) & 0.1 & 0.1990 \\
             & 0.2 & 0.2190 \\
             & 0.3 & 0.2080 \\\hline
        \end{tabular}
    }
    \label{table:asltoengbi}
\end{table}

\begin{table}[b]
    \centering
    \caption{ASL-to-ENG: Trigrams}
     \resizebox{0.95\columnwidth}{!} {
        \begin{tabular}{|c|c|c|}
        \hline
        Queue Size & Language Model Weight & BLEU-2 Score\\\hline
            8 (trigram) & 0.1 & 0.0950 \\
            & 0.2 & 0.1610 \\
            & 0.3 & 0.1610 \\\hline
            10 (trigram) & 0.1 & 0.1270 \\
            & 0.2 & 0.1440 \\
            & 0.3 & 0.1160 \\\hline
            20 (trigram) & 0.1 & 0.2760 \\
            & 0.2 & 0.1690 \\
            & 0.3 & 0.0505 \\\hline
        \end{tabular}
     }
    \label{table:asltoengtri}
\end{table}

\begin{table}[h!]
    \centering
    \caption{ENG-to-ASL Table: Language Model Weights}
    \resizebox{0.8\columnwidth}{!} {
        \begin{tabular}{|c|c|c|}
        \hline
         Queue Size & Language Model Weight & BLEU-2 Score\\\hline
           20 & 0.1 & 0.1790\\
            & 0.2 & 0.1140\\
            & 0.3 & 0.1150\\\hline
           15 & 0.1 & 0.1004\\
            & 0.2 & 0.1000\\
            & 0.3 & 0.0680\\\hline
           13 & 0.1 & 0.0964\\
            & 0.2 & 0.1480\\
            & 0.3 & 0.0860\\\hline
        \end{tabular}
    }
    \label{table:engtoasl}
\end{table}

As mentioned in the Data section, we split the dataset into three separate sets: a training set, a development set, and a test set. For both directions, we train our models on a training set, tune the hyperparameters (i.e. the language model weight, the queue size $k$, and the type of n-gram model we are using for the language model [unigram, bigram, trigram]) on a development set, and finally find the average BLEU-2 score on a test set. Furthermore, our metric for measuring the accuracy of translation on the test set or the development set is the average of individual BLEU-2 scores of the translations. Thus, any BLEU-2 score that is reported in tables or figures will be the mean BLEU-2 score.
\subsection{Hyperparameter Tuning Results}
To find the optimal hyperparamaters, we ran several experiments. The initial set of experiments were run to find the optimal hyperparameters for translation from ASL to English. First, we found the BLEU-2 score on the development set by keeping the queue size constant and adjusting the language model weight using the bigram English model. Next, we performed the same experiment except using the trigram English model. Results of these experiments are reported in tables \ref{table:asltoengbi} and \ref{table:asltoengtri}. We find that the optimal language model weight is 0.1, the optimal queue size for ASL to English translation is 20, and the optimal language model type is a trigram model. Using this combination gives us a BLEU-2 score of 0.276 (see Table \ref{table:asltoengtri}), which is the highest BLEU-2 score that we obtained while tuning hyperparameters.

To provide some intuition on the choice of experiments, we reasoned that we could converge on the optimal hyperparameter combination using a method similar to coordinate ascent, where we optimize a single parameter while keeping the other parameters constant. Then, after finding the optimal value for the parameter, repeat the process with the other parameters while still maintaining the optimal values for the parameters that have been already processed. 
Using the (20,0.1,'Trigram') combination, we ran our system on the test set; the BLEU-2 score, reported in Table \ref{table:comparison}, is 0.1202.

The subsequent experiments involved finding the optimal hyperparameters for translation from English to ASL. Using a similar methodology to the one used in the other translation direction, we ran one set of experiments where the queue size was kept constant while the language model weight was adjusted. Because we only used a unigram ASL language model while translating in this direction, running only this set of experiments is sufficient to find the optimal hyperparameter pair. Results of the experiment are reported in Table  \ref{table:engtoasl}. We find that the optimal language model weight is 0.1 and the optimal queue size is 20. Using the (20,0.1) combination and running the system on the test set, we obtain a BLEU-2 score, reported in Table \ref{table:comparison}, of 0.1802. 
\subsection{Experiments for ASL Constructions}
The last two sets of experiments test how our system performs on translating specific ASL constructions. Firstly, we perform an experiment on translating only the common comma construction described above in Modeling, by filtering for these test examples in the test set and using the optimal ASL to English hyperparameters when running on the filtered test set. The BLEU-2 score result for this test is .1201 (see Table \ref{table:comparison}). 

Finally, the last experiment tests how our system translates sign sequences with gesture tokens. We create a test set for this experiment in a similar fashion to what we did for comma constructions, by filtering the original test set to only include sign sequences with gesture tokens. The BLEU-2 score for this test is .1232 (see Table \ref{table:comparison}).

\section{Analysis}
As a sanity check for our results, it is important to note that our translation system outperforms the baseline implementations in both directions.
We also observe a difference in performance between the two translational directions.
The ASL-to-English direction produces an average BLEU-2 score of 0.12, while the English-to-ASL gives a higher score of 0.18. The behavior is consistent with our expectations going into the project since in going from English to ASL, we essentially filter out unnecessary information, whereas the other direction requires generating information in effect.

After running our tuning experiments on the hyperparameters, we noticed a few interesting trends.
First, instances in which we use a trigram model for English tend to produce higher BLEU-2 results than with corresponding bigram English models, particularly when the maximum queue size in the decoding algorithm is large. 
This makes sense because the trigram model inherently captures more information than the bigram model.
Furthermore, we see from the two graphs above that for large maximum queue sizes, instances with small language model weights perform better.
We can explain this as follows: For large maximum queue sizes, after the first few iterations of the decoding algorithm, we will see hypotheses with common words. If we do not sufficiently dampen the influence of the language model, we will begin to see hypotheses with high priorities resulting from common ngrams, and these will steer the sentence generation away from the original meaning of the sign sentence.
Therefore, we conclude that models with high maximum queue sizes and low language model weights generally perform best.

Although it is useful to look at how changing these hyperparameters changes the model, it is perhaps even more important to recognize the shortcoming of our current model.
The first issue is with our corpus, which by most measures is far too small to perform accurate machine translation.
We also lack large databases containing sign language as text which limited our language model for ASL. Unfortunately, our ASL language model was far too crude to produce accurate results. We oversimplified the model in several ways, in particular with our comma trigram structure and treatment of gestures as equivalent to other signs. Because we are not fluent in ASL, let alone fully understand its underlying structure, we had difficulty in designing its language. In fact, to our knowledge, no one else has designed a text-based language model for any sign language. Finally, machine translation for single sentences is inherently a difficult task, since the sentences lack context.
Despite these many challenges and difficulties, our system still translates many sentences effectively between the two languages in both directions. 

\section{Conclusion \& Future Work}
While we would like our translations to be as accurate as possible, it is important to note that it is virtually impossible, even for an oracle, to come up with a perfect translation. 
In any language, words and sentences carry contextual meaning that might be impossible to express exactly in another language.
The problem of translating between sign languages and natural languages is extremely difficult, and we are likely the first to address it exactly this way.
Although we faced numerous challenges, we have shown that this approach is a reasonably effective improvement to our baseline algorithms.
With a more thorough understanding of ASL and its grammatical structure, along with a larger training corpus, our approach has the potential to be an effective system of translation between ASL and English.

\section{References}
\nocite{aslgrammar}
\nocite{aslcorpus}
\nocite{asldict}
\nocite{asldict2}
\nocite{decoding}
\nocite{martinez2006generalized}
\nocite{papineni2002bleu}
\nocite{cs224n2}
\nocite{ngrams}
\nocite{cs224n}
\nocite{ibm}
\nocite{parton2006sign}
\nocite{mazzei2013deep}
\nocite{zhao2000machine}
\nocite{nltk}
\printbibliography[heading=none]
\end{document}